\documentclass[sigconf]{acmart}
\AtBeginDocument{%
  }
\usepackage[framemethod=TikZ]{mdframed}
\usepackage{multirow}
\setcopyright{acmlicensed}
\copyrightyear{2025}
\acmYear{2025}
\setcopyright{acmlicensed}\acmConference[MM '25]{Proceedings of the 33rd ACM International Conference on Multimedia}{October 27--31, 2025}{Dublin, Ireland}
\acmBooktitle{Proceedings of the 33rd ACM International Conference on Multimedia (MM '25), October 27--31, 2025, Dublin, Ireland}
\acmDOI{10.1145/3746027.3761988}
\acmISBN{979-8-4007-2035-2/2025/10}

\begin{document}

\title{Identity-Preserving Text-to-Video Generation Guided by \\ Simple yet Effective Spatial-Temporal Decoupled Representations}


\author{Yuji Wang}
\authornote{All the authors contributed equally to this research.}
\authornote{During internship at Tencent Youtu Lab.}
\affiliation{%
  \institution{Shanghai Jiao Tong
University, Tencent Youtu Lab}
  \city{Shanghai}
  \country{China}
  }
\email{yujiwang@sjtu.edu.cn}

\author{Moran Li}
\authornotemark[1]
\affiliation{%
  \institution{Tencent Youtu Lab}
  \city{Shanghai}
  \country{China}}
\email{moranli@tencent.com}

\author{Xiaobin Hu}
\authornotemark[1]
\affiliation{%
  \institution{Tencent Youtu Lab}
  \city{Shanghai}
  \country{China}}
\email{xiaobinhu@tencent.com}

\author{Ran Yi}
\authornote{Corresponding authors.}
\affiliation{%
  \institution{Shanghai Jiao Tong University}
  \city{Shanghai}
  \country{China}}
\email{ranyi@sjtu.edu.cn}

% \author{Jiangning Zhang, Han Feng, Weijian Cao, Yabiao Wang, Chengjie Wang}
% \affiliation{
%   \institution{Tencent Youtu Lab}
%   \city{Shanghai}
% \email{vtzhang@tencent.com}

\author{Jiangning Zhang}
\affiliation{
  \institution{Tencent Youtu Lab}
  \city{Shanghai}
  \country{China}}

\author{Han Feng}
\affiliation{
  \institution{Tencent Youtu Lab}
  \city{Shanghai}
  \country{China}}

\author{Weijian Cao}
\affiliation{
  \institution{Tencent}
  \city{Shanghai}
  \country{China}}

\author{Yabiao Wang}
\affiliation{%
  \institution{Tencent}
  \city{Shanghai}
  \country{China}}

\author{Chengjie Wang}
\affiliation{%
  \institution{Tencent}
  \city{Shanghai}
  \country{China}}

\author{Lizhuang Ma}
\authornotemark[3]
\affiliation{
  \institution{Shanghai Jiao Tong University}
  \city{Shanghai}
  \country{China}
}
\email{ma-lz@cs.sjtu.edu.cn}

\renewcommand{\shortauthors}{Yuji Wang et al.}

\begin{abstract}
Identity-preserving text-to-video (IPT2V) generation, which aims to create high-fidelity videos with consistent human identity, has become crucial for downstream applications. However, current end-to-end frameworks suffer a critical spatial-temporal trade-off: optimizing for spatially coherent layouts of key elements (\textit{e.g.,} character identity preservation) often compromises instruction-compliant temporal smoothness, while prioritizing dynamic realism risks disrupting the spatial coherence of visual structures. To tackle this issue, we propose a simple yet effective spatial-temporal decoupled framework that decomposes representations into spatial features for layouts and temporal features for motion dynamics. Specifically, our paper proposes a semantic prompt optimization mechanism and stage-wise decoupled generation paradigm. The former module decouples the prompt into spatial and temporal components. Aligned with the subsequent stage-wise decoupled approach, the spatial prompts guide the text-to-image (T2I) stage to generate coherent spatial features, while the temporal prompts direct the sequential image-to-video (I2V) stage to ensure motion consistency. Experimental results validate that our approach achieves excellent spatiotemporal consistency, demonstrating outstanding performance in identity preservation, text relevance, and video quality. By leveraging this simple yet robust mechanism, our algorithm secures the runner-up position in 2025 ACM Multimedia Challenge. Our code is available at \url{https://github.com/rain152/IPVG}.
\end{abstract}

%  ---- for publication version
\begin{CCSXML}
<ccs2012>
 <concept>
  <concept_id>00000000.0000000.0000000</concept_id>
  <concept_desc>Do Not Use This Code, Generate the Correct Terms for Your Paper</concept_desc>
  <concept_significance>500</concept_significance>
 </concept>
 <concept>
  <concept_id>00000000.00000000.00000000</concept_id>
  <concept_desc>Do Not Use This Code, Generate the Correct Terms for Your Paper</concept_desc>
  <concept_significance>300</concept_significance>
 </concept>
 <concept>
  <concept_id>00000000.00000000.00000000</concept_id>
  <concept_desc>Do Not Use This Code, Generate the Correct Terms for Your Paper</concept_desc>
  <concept_significance>100</concept_significance>
 </concept>
 <concept>
  <concept_id>00000000.00000000.00000000</concept_id>
  <concept_desc>Do Not Use This Code, Generate the Correct Terms for Your Paper</concept_desc>
  <concept_significance>100</concept_significance>
 </concept>
</ccs2012>
\end{CCSXML}

\ccsdesc[500]{Computing methodologies~Image and video acquisition}

\keywords{Video Generation; Diffusion Transformer; AIGC; Multi-Modal Generation; Prompt Optimization}
\begin{teaserfigure}
  \setlength{\abovecaptionskip}{0pt}
  \centering
  \includegraphics[width=0.95\textwidth]{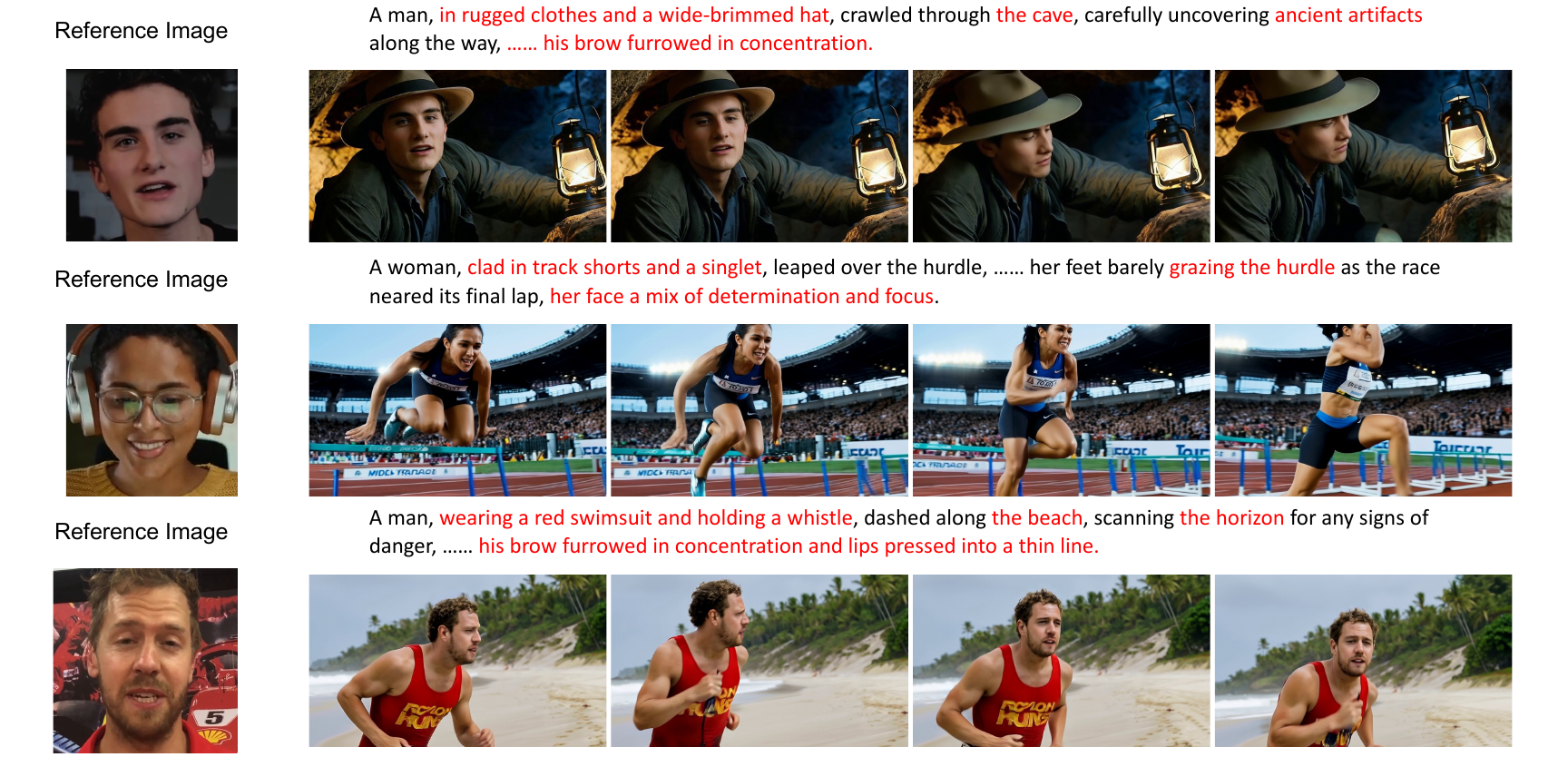}
  \caption{Examples of identity-preserving text-to-video generation (IPT2V) by our stage-wise decoupled framework. Given a reference image,  it can generate high-fidelity human videos with rich details and consistent identity. {\color{red} Red} text highlights key elements in long instructions. All facial images are from the VIP-200K test set~\cite{vip-200k}.}
  \label{fig:teaser}
\end{teaserfigure}

%% information and the body of the document, and typically spans the page. 

\maketitle
\section{Introduction}
Recent breakthroughs in diffusion-based video generation ~\cite{ho2022video,zheng2024open,yang2024cogvideox,xing2024survey,hu2024motionmaster} have enabled the creation of highly realistic videos with smooth temporal continuity. 
% In critical applications like virtual influencer marketing, 
In critical applications such as animation or advertising production~\cite{chen2025identity,xiaobin2025vtbench,lu2024pinco,tang2025ata}, maintaining consistent visual identities across frames has become a core requirement. This demand has spurred targeted research on identity-preserving text-to-video (IPT2V) ~\cite{chefer2024still,wei2025echovideo,zhang2025motion},  which aims to generate text-aligned videos while preserving precise human identities.

Many recent studies have focused on the IPT2V task, with models like VACE~\cite{jiang2025vace}, SkyReels-A2~\cite{fei2025skyreels}, and HunyuanCustom~\cite{hu2025hunyuancustom} demonstrating remarkable capabilities through large-scale training and advanced architectures. Current end-to-end frameworks face an inherent spatial-temporal trade-off: optimizing for consistent key element layouts (\textit{e.g.,} preserving character identity across frames) often harms instruction-aligned smooth motion, while focusing on dynamic realism can break visual structure consistency in facial features or clothing. The challenge lies in that joint optimization of these conflicting objectives inherently complicates the balance between spatial coherence and temporal realism. Additionally, current approaches rarely use prompt guidance for spatial-temporal disentanglement, leading to inconsistent identities and motions. 

To address the challenges above, we propose a simple yet effective spatial-temporal decoupled framework with semantic optimization. First, we propose a stage-wise decoupled generation paradigm which decouples IPT2V into two stages: a text-to-image (T2I) stage that focuses on high-fidelity spatial identities and semantically aligned key visual element layouts and an image-to-video (I2V) stage for temporal modeling to ensure motion smoothness and spatiotemporal coherence. Second, we propose a semantic instruction optimization mechanism alongside the framework to decompose prompts into spatial-temporal decoupled representations, which consists of two modules: the Spatial Semantic Parser extracts static semantics to anchor T2I identity modeling, while the Temporal Prompt Polisher optimizes sentence logic and enhances dynamism for I2V guidance. This design separates layout and motion cues to bridge text-semantic gaps for coherent video generation. 

Our contribution can be summarized as follows:

\begin{itemize}
\item We propose a simple yet effective spatial-temporal decoupled framework for IPT2V, leveraging the stage-wise decoupled generation paradigm. In this framework, T2I enables high-fidelity spatial layout modeling and I2V optimizes temporal consistency. This approach resolves the spatial-temporal conflict in end-to-end T2V frameworks and boosts efficiency. 

\item We propose two specialized modules for semantic prompt optimization: the Spatial Semantic Parser assists T2I identity modeling by extracting static spatial semantics, while the Temporal Prompt Refiner enhances I2V motion realism through logical refinement and dynamic cue addition. 

\item Experiments show that our method demonstrates superior performance in identity preservation, text relevance, and video quality, securing the runner-up in 2025 ACM Multimedia Challenge. This verifies the simplicity and effectiveness of our spatial-temporal decoupled framework.
\end{itemize}
\begin{figure*}[t]
    \centering
    \includegraphics[width=0.95\textwidth]{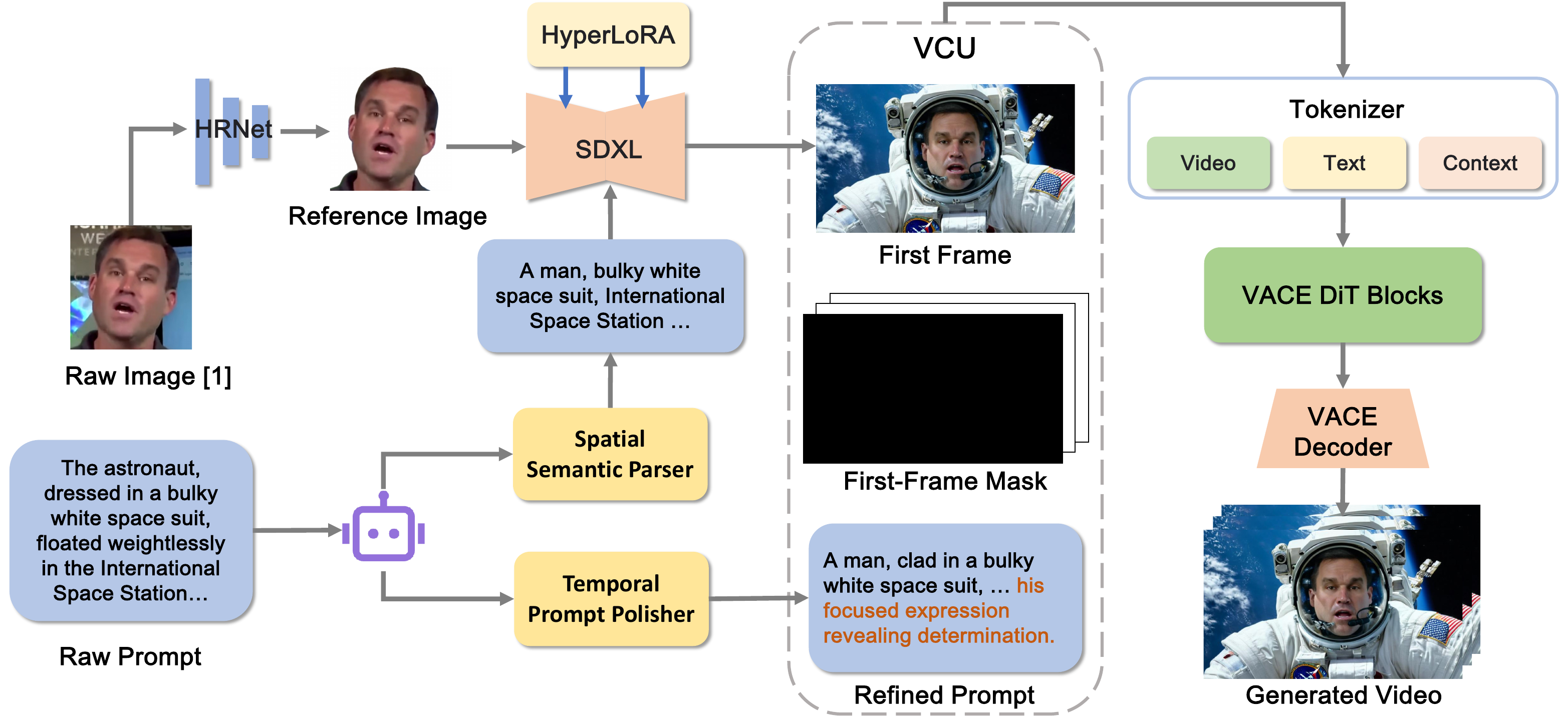}
    \vspace{-2mm}
    \caption{The pipeline of our proposed framework, comprising a spatial-temporal decoupling architecture. In the T2I stage, a pre-trained HRNet~\cite{wu2021optimized} generates a human-centric Reference Image via background removal, while a Spatial Semantic Parser extracts character-relevant phrases from input prompts. These are fed into HyperLoRA to produce a high-fidelity first-frame image. The I2V stage employs a Temporal Prompt Polisher to refine instructions for logical refinement and dynamic enhancement, followed by VACE-based~\cite{jiang2025vace} first-frame generation to synthesize temporally coherent videos.} 
    \label{fig:overview}
    \vspace{-3mm}
\end{figure*}

\section{Related Work}
\textbf{Identity-Preserving Text-to-Video Generation.} 
For IPT2V, early approaches primarily relied on single-case fine-tuning~\cite{chefer2024still,wu2024motionbooth,wei2024dreamvideo,zhong2025concat}, which proves excessively time-consuming. Consequently, tuning-free research has emerged as the dominant direction. Among these, MovieGen~\cite{ehtesham2025movie} and ID-Animator~\cite{he2024idanimator} pioneered IPT2V generation. MovieGen is closed-source, and ID-Animator uses image-model methods that compromise identity consistency in dynamic frames. ConsisID~\cite{yuan2025consisid} uses facial frequency signals but suffers pose inconsistencies from 2D representations, while  FantasyID~\cite{zhang2025fantasyid} addresses this with 3D facial priors. Recently, VACE~\cite{jiang2025vace}, SkyReels-A2~\cite{fei2025skyreels} augment Wan2.1~\cite{wan2025wan} with reference image branches to enable person-specific generation. However, end-to-end joint optimization inherently leads to a spatial-temporal trade-off, undermining layout coherence and facial detail fidelity during dynamic motion synthesis. To resolve this conflict, we propose a stage-wise decoupled paradigm: T2I for static spatial key-element layout modeling, and I2V for dynamic temporal motion. 

\noindent \textbf{Text-to-image Portrait Synthesis.} Text-to-image (T2I) synthesis has evolved rapidly from early transformer/GAN architectures~\cite{yu2022scaling,kang2023scaling,goodfellow2014generative,hu2020face,hu2022autogan} to diffusion-based pipelines~\cite{dhariwal2021diffusion,ho2022cascaded, hu2024diffumatting}. Unlike early methods hampered by limited capacity and poor text-image alignment, diffusion models now excel in generating high-fidelity images for datasets with large intra-domain variance~\cite{schuhmann2021laion}. For human-centered portrait generation, some tuning-free methods, such as PhotoMaker~\cite{li2024photomaker}, T2I-Adapter~\cite{mou2024t2i}, IP-Adapter \cite{cui2024idadapter}, and InstantID~\cite{wang2024instantid}, address efficiency by introducing plug-in structures that encode identity features without retraining. Recent work HyperLoRA~\cite{li2025hyperlora} further decouples facial information, enabling stronger capabilities in character fidelity. T2I has better control over image layout and facial details.
In our framework, to further enhance generation quality, we introduce a Spatial Semantic Parser that extracts critical semantic entities from input prompts, guiding the T2I stage to prioritize key-element layout coherence.

\noindent \textbf{Human-Centric Image-to-Video Generation.}
Image-to-video (I2V) generation faces challenges in steering content with text alone, prompting recent methods to integrate first-frame conditioning for controllability. Approaches such as AnimateAnything~\cite{dai2023animateanything}, and PixelDance~\cite{zeng2024make} modify T2V U-Net architectures by concatenating first-frame latent features with input noise to enable temporal consistency via spatial feature anchoring. Dynamicrafter~\cite{xing2024dynamicrafter}, and Moonshot~\cite{zhang2024moonshot} further enhance conditioning by incorporating image cross-attention layers to inject stronger visual cues into the diffusion process. Recently, works such as Wan2.1~\cite{wan2025wan}, VACE~\cite{jiang2025vace}, and FramePack~\cite{zhang2025packing} also support I2V tasks, demonstrating outstanding performance through extensive training datasets and intricate model architecture design. To enhance I2V generation, we propose a Temporal Prompt Polisher to boost video temporal dynamics by enforcing logical coherence and injecting dynamic motion cues. 

\footnotetext[1]{The raw image in Fig. \ref{fig:overview} comes from VIP-200K test set~\cite{vip-200k}.}
\section{Method}

\subsection{Overview}
We propose a novel framework to address the IPT2V problem, integrating a stage-wise decoupled generation paradigm and semantic prompt optimization. Traditional end-to-end IPT2V models struggle to maintain coherent spatial layouts of key elements and temporal motion dynamism effectively. To tackle this, we introduce a spatial-temporal decoupled framework for IPT2V, where T2I technology enables high-fidelity spatial layout modeling, and I2V technology optimizes motion smoothness and dynamism. This paradigm resolves the spatial-temporal conflict in end-to-end T2V and achieves an efficient generation process. Additionally, to complement our framework, we propose semantic prompt optimization to decompose prompts into spatial-temporal decoupled representations where a Spatial Semantic Parser extracts critical visual entities and a Temporal Prompt Polisher enhances I2V motion realism through logical refinement and dynamic cue addition. 

The pipeline of our proposed architecture, as illustrated in Fig. \ref{fig:overview}, unfolds in two sequential stages. In the T2I stage, the Spatial Semantic Parser extracts critical entities from raw instructions, which are merged with an HRNet-preprocessed reference face image. This combined input is fed into HyperLoRA to generate the first-frame image. In the I2V stage, a Temporal Prompt Polisher optimizes the original prompt to derive a refined version that emphasizes character details and dynamic motion cues. The first-frame image is then input into our VACE-based image-to-video model, enabling the generation of spatiotemporally coherent video sequences. 

\subsection{High-Fidelity First Frame Generation}
In the T2I stage, we generate the first-frame image from face images and prompt semantics to facilitate I2V.  We adopt HyperLoRA~\cite{li2025hyperlora} as the backbone for its advantages: decoupled ID-LoRA/Base-LoRA enables parameter-efficient high-fidelity face modeling, and zero-shot inference adapts to diverse prompts without fine-tuning. This makes HyperLoRA ideal for generating high-fidelity first frames.

In image preprocessing, the original reference image's background hinders facial feature extraction. Considering that segmentation preprocessing \cite{hu2023high} facilitates the disentanglement between identity (ID) and non-identity (non-ID) priors, we use HRNet~\cite{wu2021optimized} for background removal to highlight contours, textures, and expressions, enabling the model to capture fine-grained facial characteristics more precisely. 
For prompt optimization, we propose the Spatial Semantic Parser module, built on LLMs to extract key visual elements with a focus on facial close-ups, clothing, and environments. In addition, we apply two filtering rules: 1) excluding irrelevant details (e.g., shoes/feet) that are unrelated to facial features and disrupt layout coherence; 2) removing background characters, as they may trigger unintended facial feature injection and degrade identity purity. Since HyperLoRA relies on core semantic phrases, the parser refines prompts to prioritize facial prominence, maximizing HyperLoRA’s T2I potential. The instruction template of the Spatial Semantic Parser is as follows:
\begin{figure}[h]
    \vspace{-2mm}
    \centering 
    \includegraphics[width=\linewidth]{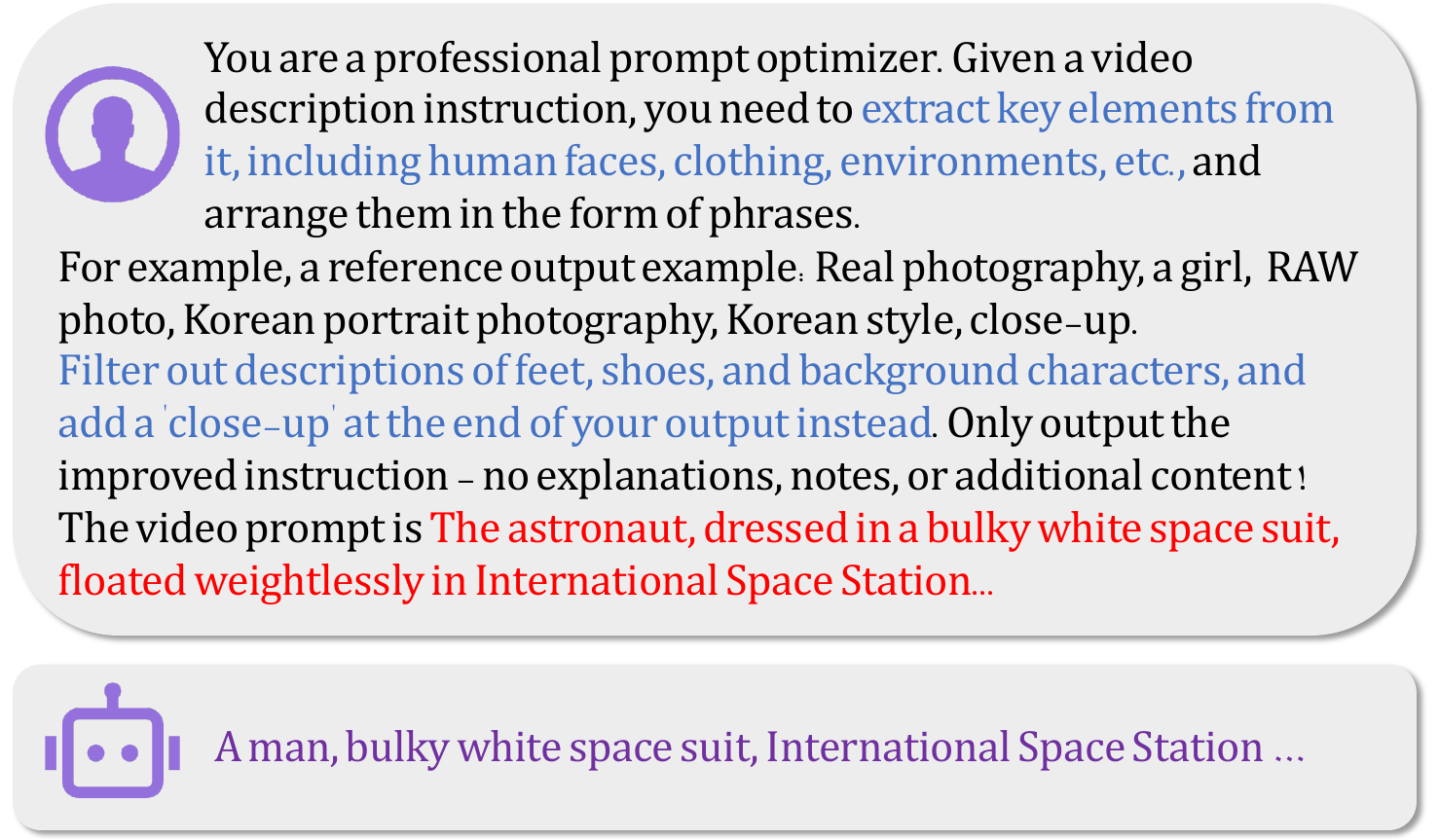}
    \label{fig:template1}
    \vspace{-6mm}
\end{figure}

The whole T2I stage can be formulated as:
\begin{equation}
I_{\text{first}} = \text{SDXL}\left( f_{\text{HR}}(I_{\text{ref}}), P_\text{spatial}; \theta_{\text{ID}}, \theta_{\text{Base}} \right),
\end{equation}
where \(I_{\text{first}}\) is the generated first-frame image, \(f_{\text{HR}}\) denotes HRNet, \(P_{\text{spatial}}\) is the optimized spatial semantic prompt, \(\theta_{\text{ID}}\)  and \(\theta_{\text{Base}}\) refer to HyperLoRA weights.

\subsection{Image-to-Video Generation with Temporal Prompt Polishment}

In the I2V stage, we generate the corresponding video by combining the first-frame image with the prompts. Considering VACE~\cite{jiang2025vace} harnesses a diverse and comprehensive training dataset and features a well-designed Video Condition Unit (VCU) that unifies multimodal inputs, here we choose VACE~\cite{jiang2025vace} as our backbone.

In the I2V task, VACE leverages the first-frame reference through its Video Condition Unit (VCU) and Concept Decoupling mechanism. The first-frame image \(I_\text{first}\) is inserted as the prefix of the frame sequence $F$, structured as: \(F = \{I_\text{first}\} + \{0_{h \times w}\} \times n\), where \(0_{h \times w}\) denotes an empty frame, and $n$ is the number of subsequent frames. The corresponding first-frame mask sequence $M$ is defined as: \(M = \{0_{h \times w}\} + \{1_{h \times w}\} \times n\), with \(0_{h \times w}\) preserving the first-frame region and\(1_{h \times w}\) enabling generation for the rest. VCU unifies inputs as \(V = [T; F; M]\) where $T$ denotes text prompts, then the Context Adapter injects these features into the Diffusion Transformer via spatiotemporal representations, ensuring first-frame consistency in generated videos. 

To facilitate instruction-guided video generation, we propose the Temporal Prompt Polisher, which enhances temporal motion realism in I2V generation through logical refinement and dynamic cue injection. Specifically, we first predict the gender of the subject in the reference image to establish explicit semantic binding with visual features. Then we restructure the prompt's logical syntax to emphasize the core actions and details of the primary subject. Meanwhile, context-aware dynamic facial cues are added to enhance temporal realism. This approach achieves motion realism by aligning semantic instructions with visual dynamics. The instruction template of the Temporal Prompt Polisher is as follows: 
\begin{figure}[h]
\vspace{-5mm}
\centering
\includegraphics[width=\linewidth]{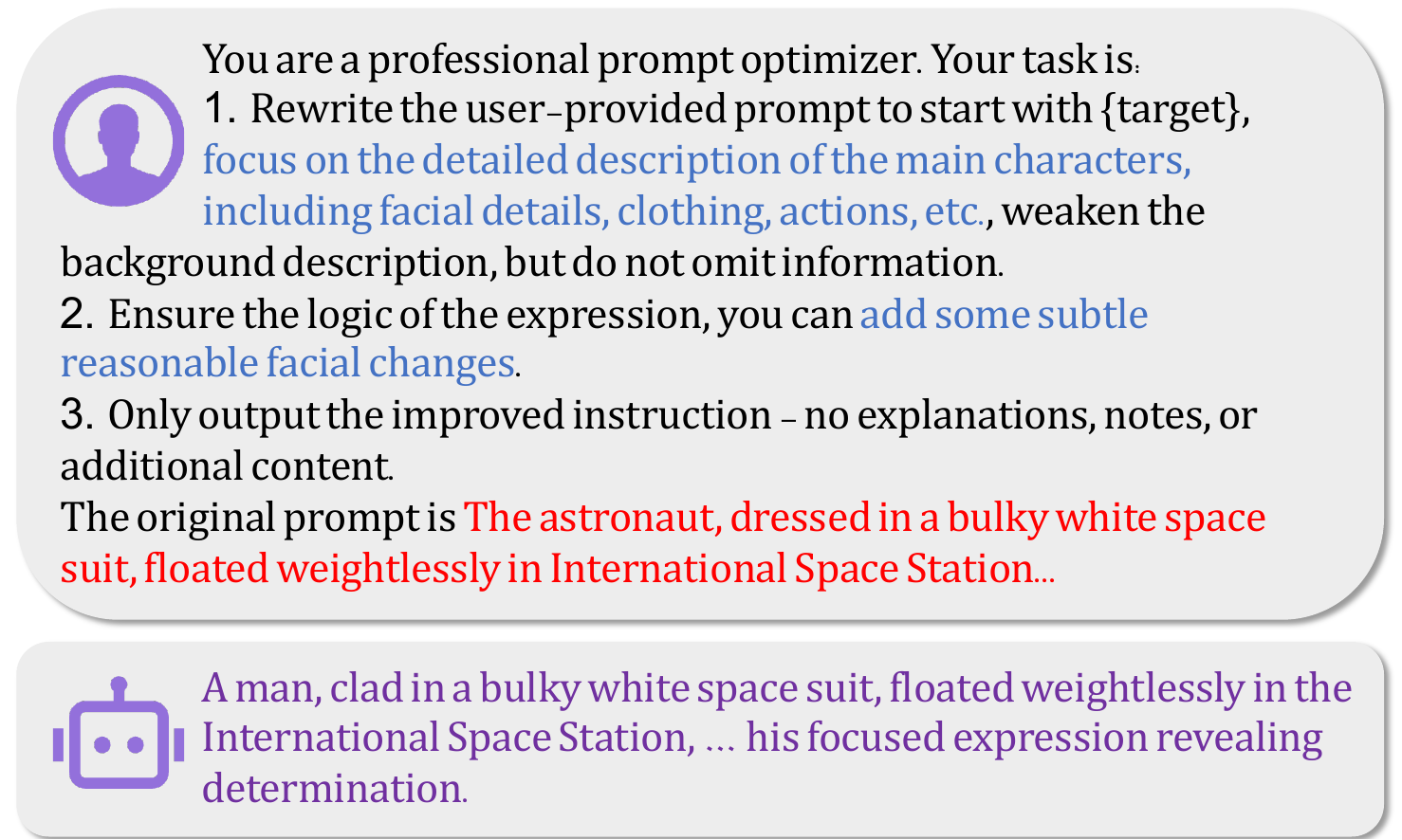}
    \label{fig:template2}
    \vspace{-6mm}
\end{figure}

The whole I2V stage can be formulated as:
\begin{equation}
V_{gen} = \text{VACE}\left( V = \left[ P_{\text{temporal}}; F; M \right];~\theta_{\text{VACE}} \right)
\end{equation}
where \(V_{gen}\) refers to generated videos, \(P_{\text{temporal}}\) denotes the temporally optimized prompt and $\theta_{\text{VACE}}$ refers to pretrained weights.

\section{Experiments}
\begin{table*}[tbp]
\renewcommand{\tabcolsep}{3pt}
\centering
\caption{Quantitative comparison on VIP-200K test set~\cite{vip-200k}. Better metrics are bolded. Our T2I→I2V pipeline outperforms R2V in identity preservation, text relevance, and key video quality metrics. Notably, it ranked 2nd among 18 teams in 2025 ACM Multimedia Challenge, far surpassing the 3rd-place Wislab particularly in identity preservation.}
\vspace{-3mm}
\begin{tabular}{lcccp{0.8cm}p{0.8cm}p{0.8cm}p{0.8cm}}
\hline
\multirow{2}{*}{Pipeline}    & \multicolumn{2}{c}{Identity Preservation}  & Text Relevance & \multicolumn{4}{c}{Video Quality}          \\ \cline{2-3} \cline{5-8}
 & FaceSim-Arc↑ & FaceSim-Cur↑ & CLIP-Score↑ & AQ↑ & IQ↑ & MS↑ & DD↑ \\ \hline
Wislab & 0.269 & 0.285 & \textbf{28.600} & \multicolumn{1}{c}{-} & \multicolumn{1}{c}{-} & \multicolumn{1}{c}{-} & \multicolumn{1}{c}{-} \\
R2V      & 0.406 & 0.423 & 27.868 & 0.583 & 0.692 & \textbf{0.984} & 0.552 \\
T2I→I2V (Ours) & \textbf{0.441} & \textbf{0.467} & 27.984 & \textbf{0.588} & \textbf{0.701} & 0.979 & \textbf{0.848}\\
\hline
\end{tabular}
\label{tab:1}
\vspace{-2mm}
\end{table*}
\begin{figure*}[t]
    \centering
    \includegraphics[width=\textwidth]{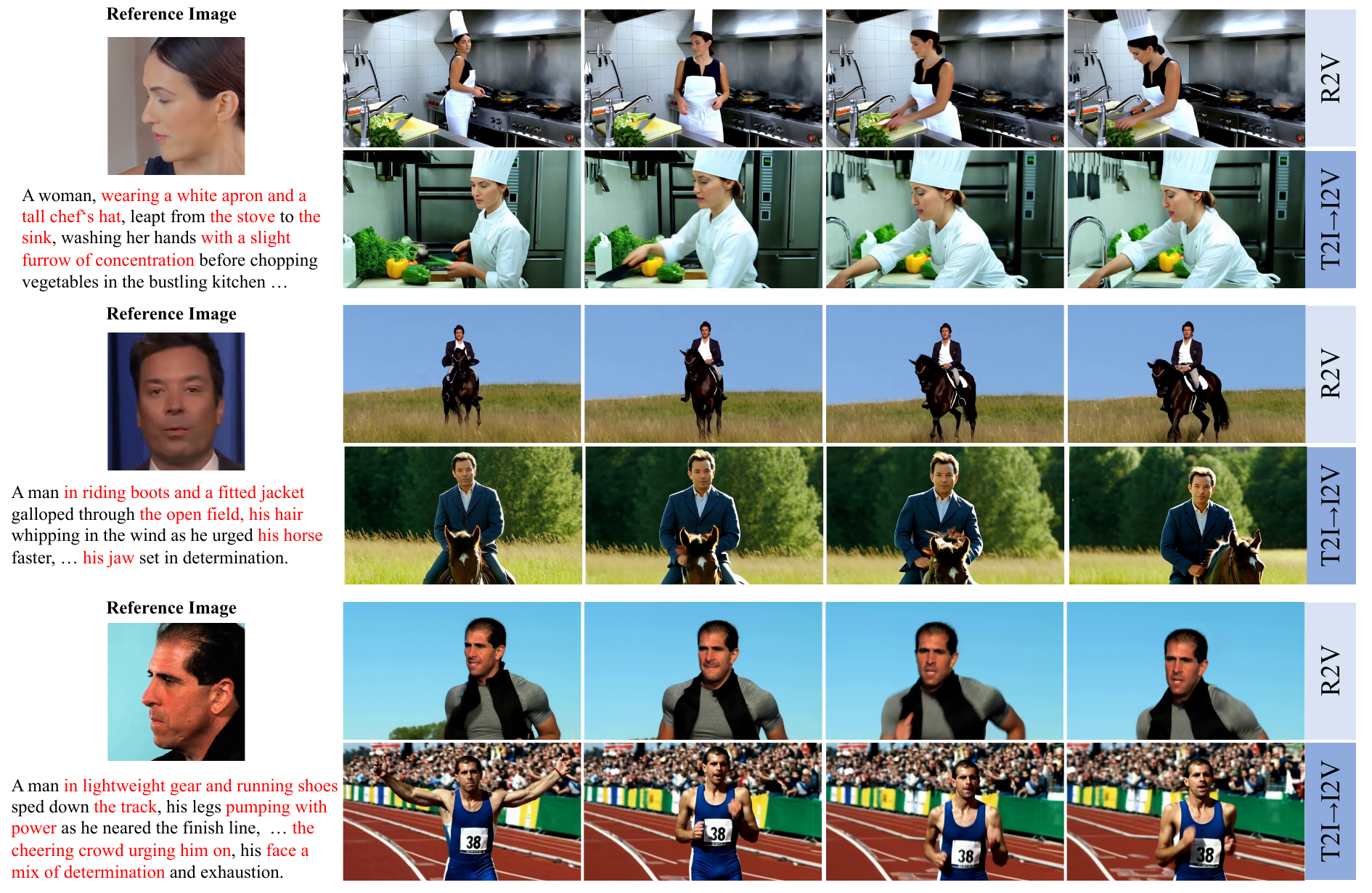}
    \vspace{-6mm}
    \caption{Qualitative analysis between R2V and T2I+I2V pipeline. The T2I+I2V framework demonstrates better ID preservation capability, and the video content aligns more effectively with the text. All facial images are from the VIP-200K test set~\cite{vip-200k}.}
    \label{fig:exp}
\end{figure*}

\subsection{Implementation Details}
\noindent \textbf{Models.} We adopt Qwen3-8B~\cite{yang2025qwen3} as the foundation model for the Spatial Semantic Parser and Temporal Prompt Polisher. For T2I tasks, HyperLoRA-T2I-FaceDetailer generates first-frame images with high-fidelity facial features, while VACE-14B serves as the backbone for I2V, enabling coherent temporal generation via VCU.

\noindent \textbf{Metrics.} To comprehensively evaluate video generation quality, we explore three aspects: identity preservation, text relevance, and video quality. Following ConsisID~\cite{yuan2025consisid}, we select FaceSim-Arc~\cite{huang2020curricularface} and FaceSim-Cur~\cite{deng2019arcface} as metrics to measure identity preservation, both indicators calculate the similarity between the extracted facial features of each video frame and those of the reference image. For text relevance, we adopt CLIP-Score to assess the alignment between generated video content and input textual prompts. For video quality, we select several metrics.  Aesthetic Quality (AQ) is used to calculate the aesthetic score of a video, which is evaluated by the LAION model~\cite{LAION-AI2022aesthetic-predictor}. Imaging Quality (IQ) is evaluated with the MUSIQ~\cite{ke2021musiq} model, assessing low-level distortions (e.g., over-exposure, noise, blur). For Motion Smoothness (MS), we compute MAE between original odd frames and those reconstructed by a video interpolation model AMT~\cite{li2023amt} after dropping odd frames. Dynamic Degree (DD) is defined as the proportion of non-static videos, measured by averaging the top 5\% optical flow strengths via RAFT~\cite{teed2020raft}.

\begin{table*}[tbp]
\renewcommand{\tabcolsep}{3pt}
\centering
% \vspace{-5mm}
\caption{Ablation study of the Role-Focused Prompt Polisher. Better metrics are indicated by bolding. With the same first-frame driving, videos generated by the improved prompts exhibit better quality than those by raw prompts, outperforming in all evaluated metrics. This pattern holds for both VACE-14B~\cite{jiang2025vace} and Wan-1.3B~\cite{wan2025wan}, verifying the module's effectiveness.}
\vspace{-2mm}
\begin{tabular}{lccccp{0.8cm}p{0.8cm}p{0.8cm}p{0.8cm}}
\hline
\multirow{2}{*}{Backbone} & \multirow{2}{*}{Prompt} & \multicolumn{2}{c}{Identity Preservation}  & Text Relevance & \multicolumn{4}{c}{Video Quality}          \\ \cline{3-4} \cline{6-9}
& & FaceSim-Arc↑ & FaceSim-Cur↑ & CLIP-Score↑ & AQ↑ & IQ↑ & MS↑ & DD↑ \\ \hline
 \multirow{2}{*}{VACE-14B} & Raw   & 0.395 & 0.410 & 27.835 & 0.579 & 0.680 & \textbf{0.976} & 0.776 \\
 & Refined & \textbf{0.436} & \textbf{0.462} & \textbf{27.943} & \textbf{0.586} & \textbf{0.689} & \textbf{0.976} & \textbf{0.812} \\
 \hline
\multirow{2}{*}{Wan-1.3B} & Raw & 0.270 & 0.286 & 27.374 & 0.520 & 0.665 & 0.964 & 0.824 \\
& Refined & \textbf{0.320} & \textbf{0.338} & \textbf{27.463} & \textbf{0.535} & \textbf{0.679} & \textbf{0.968} & \textbf{0.840} \\
\hline
\end{tabular}
\label{tab:abla}
\end{table*}

\begin{figure}[t]
    \centering
    \includegraphics[width=\linewidth]{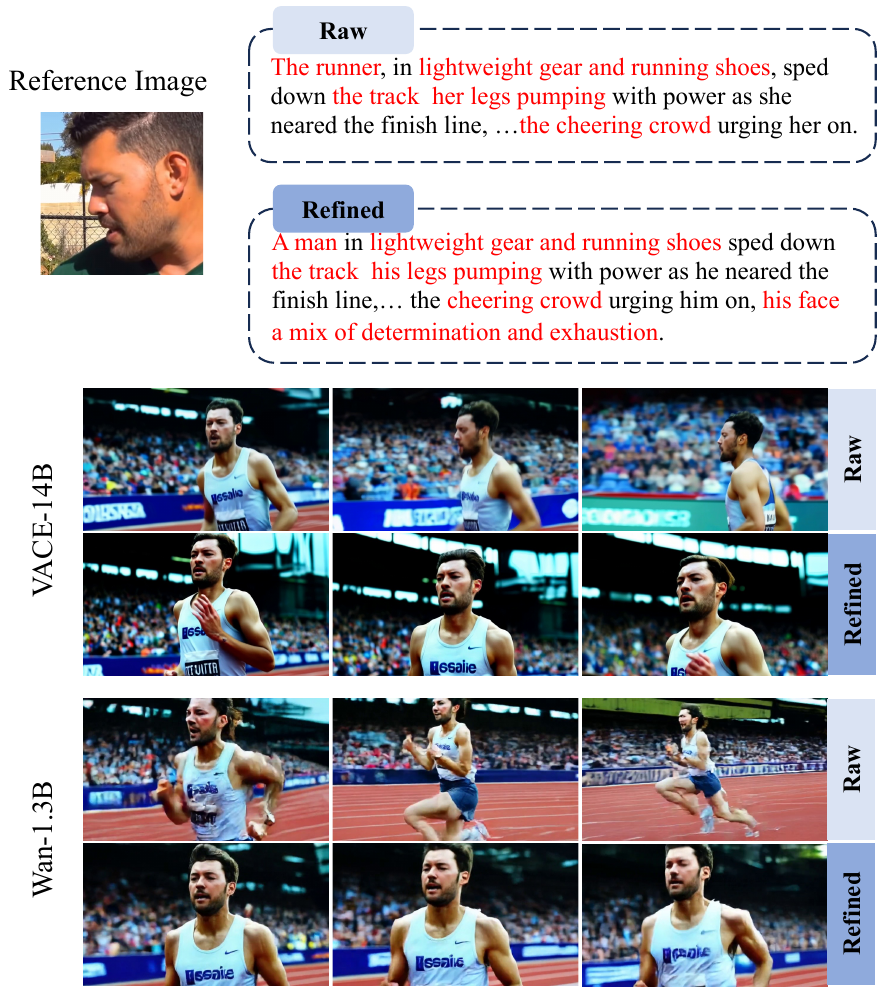}
    \vspace{-4mm}
    \caption{Qualitative analysis of the ablation study. Results show that videos driven by refined prompts have better facial details and higher quality. All facial images are from the VIP-200K test set~\cite{vip-200k}.}
    \label{fig:abla}
    \vspace{-4mm}
\end{figure}
\subsection{Quantitative and Qualitative Analysis}
\label{main_exp}
In this section, we present the quantitative results and qualitative analysis of the proposed framework on the VIP-200K test set~\cite{vip-200k}. For the baseline selection, we use the Reference-to-Video (R2V) method of VACE-14B, which directly uses facial images as a reference to generate videos, for comparison. 

Table \ref{tab:1} shows our T2I→I2V pipeline outperforms end-to-end R2V in identity preservation, text relevance, and video quality. Specifically, Our approach achieves an FaceSim-Cur score of 0.467 (10.4\% improvement over R2V's 0.423), confirming the decoupled T2I stage effectively anchors facial features. For text relevance, the CLIP-Score of our approach is 27.984,  higher than R2V's 27.868, verifying better alignment with textual instructions. In video quality, the dynamic score of T2I→I2V is notably superior at 0.848, representing a 53.6\% increase compared to R2V's 0.552, indicating more natural video motion. Although the smoothness metric is slightly lower (0.979 vs. 0.984), the significant boost in dynamic performance demonstrates that T2I→I2V achieves a better balance between motion richness and temporal consistency. These results demonstrate the superior generation effects of our T2I→I2V pipeline. 

Fig. \ref{fig:exp} presents comparative visual examples.  The decoupled T2I→I2V framework preserves identities better with clearer facial details, and generates smoother motions with stronger text alignment. For instance, in the third example, the video generated by R2V includes only one character, losing both the crowd and track elements described in the text, missing fine-grained actions, while T2I→I2V retains scene complexity and natural motions that match descriptions.
% while failing to capture the fine details of the character's movements. In contrast, T2I→I2V excels in content layout, simultaneously ensuring smoother and more natural character motions, which faithfully align with the textual descriptions.

\subsection{Ablation Study}
\textbf{Effects of the Spatial-Temporal Decoupled Representations.} As analyzed in Sec. \ref{main_exp}, the decoupled T2I→I2V framework achieves superior performance by separating spatial feature anchoring (via the first-frame generation) and temporal motion synthesis. The T2I stage provides explicit spatial layout (object positioning, scene composition) to guide generator spatial consistency, reducing motion ambiguity and enhancing video rationality. 

\noindent \textbf{Effects of the Temporal Prompt Polisher.} In the I2V stage of our framework, we propose a Temporal Prompt Polisher to refine instructions. To validate its effectiveness, ablation experiments on the I2V task (first-frame generated by HyperLoRA) compare raw and refined prompts using VACE-14B and Wan-1.3B to evaluate cross-scale applicability. We select 25 diverse identities from the test dataset for comprehensive validation. 

Table \ref{tab:abla} presents ablation results for the Role-Focused Prompt Polisher, showing that refined prompts improve I2V performance across both VACE-14B and Wan-1.3B models. For example, VACE-14B's FaceSim-Cur score increases from 0.410 to 0.462, highlighting enhanced facial detail representation. Consistent improvements in text relevance and video quality metrics further validate the module's efficacy. These quantitative findings are complemented by the visual insights in Fig. \ref{fig:abla}. The refined prompt corrects illogical elements in the original sentence, such as changing "her" to "his", and adds appropriate facial descriptions. Videos driven by the refined prompt exhibit superior facial details and more natural dynamics, validating the effectiveness of the Temporal Prompt Polisher. 

\section{Conclusion}
In this work, we address the challenges of identity preservation and semantic alignment in video generation by introducing a stage-wise decoupled generation paradigm and a complementary semantic prompt optimization mechanism. The framework resolves the spatial-temporal conflict by decomposing IPT2V into T2I for spatially coherent identity modeling and I2V for temporally consistent motion generation. In the semantic optimization process, our proposed Spatial Semantic Parser extracts static scene elements to align with the T2I stage, while the Temporal Prompt Polisher enhances sequential motion guidance for I2V, ensuring cross-stage semantic consistency between spatial anchoring and temporal generation. Experimental results on the VIP200k dataset demonstrate the framework's superiority, significantly improving identity preservation, text-video alignment, and motion realism. Our approach provides a simple yet effective solution for IPT2V tasks. Future research will explore cross-task feature transfer to enhance temporal coherence. 
\section*{Acknowledgements}
This work was supported by National Natural Science Foundation of China (No. 72192821, 62302297, 62472282), Young Elite Scientists Sponsorship Program by CAST (2022QNRC001), the Fundamental Research Funds for the Central Universities (project number: YG2023QNA35).
\bibliographystyle{ACM-Reference-Format}
\bibliography{sample-base}

\appendix

\end{document}